\documentclass[conference]{IEEEtran}
\usepackage{cite}
\usepackage{amsmath,amssymb,amsfonts}
\usepackage{algorithmic}

\usepackage[pdftex]{graphicx, xcolor}

\usepackage{textcomp}
\usepackage{xcolor}

\usepackage{siunitx}
\usepackage[subrefformat=parens]{subcaption}
\usepackage{booktabs}
\usepackage{multirow}
\usepackage{bm}
\def\BibTeX{{\rm B\kern-.05em{\sc i\kern-.025em b}\kern-.08em
    T\kern-.1667em\lower.7ex\hbox{E}\kern-.125emX}}
\begin{document}

\title{Handwriting Prediction Considering\\
Inter-Class Bifurcation Structures}

\author{

\IEEEauthorblockN{Masaki Yamagata, Hideaki Hayashi, and Seiichi Uchida}
\IEEEauthorblockA{\textit{Department of Advanced Information Technology} \\
\textit{Kyushu University}\\
Fukuoka, Japan \\
masaki.yamagata@human.ait.kyushu-u.ac.jp, \{hayashi,uchida\}@ait.kyushu-u.ac.jp}
}

\maketitle

\begin{abstract}
Temporal prediction is a still difficult task due to the chaotic behavior, non-Markovian characteristics, and non-stationary noise of temporal signals. Handwriting prediction is also challenging because of uncertainty arising from inter-class bifurcation structures, in addition to the above problems. For example, the classes `0' and `6' are very similar in terms of their beginning parts; therefore it is nearly impossible to predict their subsequent parts from the beginning part. In other words, `0' and `6' have a bifurcation structure due to ambiguity between classes, and we cannot make a long-term prediction in this context. In this paper, we propose a temporal prediction model that can deal with this bifurcation structure. Specifically, the proposed model learns the bifurcation structure explicitly as a Gaussian mixture model (GMM) for each class as well as the posterior probability of the classes. The final result of prediction is represented as the weighted sum of GMMs using the class probabilities as weights. When multiple classes have large weights, the model can handle a bifurcation and thus avoid an inaccurate prediction. The proposed model is formulated as a neural network including long short-term memories and is thus trained in an end-to-end manner. The proposed model was evaluated on the UNIPEN online handwritten character dataset, and the results show that the model can catch and deal with the bifurcation structures.
\end{abstract}


{\allowdisplaybreaks
\section{Introduction}
Temporal prediction is an important and still challenging task in various applications~\cite{uncertainty_work2, temporal-prediction1,temporal-prediction2, temporal-prediction3}. This difficulty comes from various reasons that cause uncertainty. Many attempts have been made in the literature to improve the accuracy of prediction by appropriately modeling the uncertainty~\cite{uncertainty_work1, uncertainty_work2}. 
\par
Despite those attempts, the perfect prediction is theoretically impossible when the target temporal signals have {\em bifurcation}. This is 
the case that there are two (or more) very different distributions in the future and we are not sure which distribution will be taken. 
As a simple but informative example, let us assume the trajectories of writing digits `0' and `6.' Their beginning parts are nearly identical, and therefore it is impossible to predict which of two will be subsequently written at their beginning part. This example suggests that perfect handwriting prediction is theoretically impossible.\par
Even though the perfect prediction is still impossible, a prediction model that can learn the underlying bifurcation structure 
automatically is very useful from several aspects. First, we can give an accurate prediction result {\em until} the bifurcation point. Second, we can generate multiple prediction results {\em beyond} the bifurcation point, if necessary. Third, if we can know there is no bifurcation after a certain point, we can determine a unique prediction result with high confidence.\par
In this paper, we propose a class-guided prediction (CGP) model and apply it to a handwriting digit prediction task. In this application, ``class'' means ten digit classes from `0' to `9.' By incorporating the class information during the training the prediction model in an explicit way, we can build a model which deals with the inter-class bifurcation, such as the above example of `0' and `6.' \par
Roughly speaking, our CGP model is derived by the factorization of the prediction task into a coordinate prediction module by class-wise GMMs and a class probability module. These modules are realized as neural networks and trained simultaneously in an end-to-end manner. The outputs of those modules are the parameters of GMMs and the class probability distribution. By using those outputs, we can provide not only the prediction result but also the uncertainty degree by the inter-class bifurcation.\par
Note that handwriting digit trajectories are simple but very suitable for observing the prediction performance of the model, although our prediction model can be applied to any temporal patterns. Especially, handwriting digit trajectories have the ten predefined classes, finite temporal lengths, and several bifurcation structures. Even from an application viewpoint, it is still useful to examine the possibility of realizing {\em early classification}; if the model tells that there is no inter-class bifurcation at the current point, we can determine the recognition result by the class with the highest probability without waiting for the end of the pattern.
\par
Our main contributions are summarised as follows:
\begin{itemize}
    \item  To the best of our knowledge, this is the first study that applies a class-guided prediction model to handwriting trajectories in the presence of ambiguities by, especially, inter-class bifurcations. 
    \item  The proposed model can provide a probabilistic prediction while evaluating uncertainty by inter-class bifurcation. This ability is useful for the future applications of the proposed model such as the early classification.
    \item The proposed model gives an end-to-end training framework for individual class probability estimation and trajectory prediction. 
\end{itemize}

\section{Related work}
\subsection{Models for online handwriting generation}
Studies on the generation of online handwriting can be divided into two approaches: One is based on the motor model and the other on the stochastic model. In the motor model-based approach, the movement of the hand while writing is formulated using differential equations based on the kinematic theory. For example, Plamondon and Maarse regarded the generation of handwriting as a product of motor behavior and expressed the writing process using a hand movement model~\cite{motion-model1}. Plamondon and Guerfali proposed a model of handwriting generation that parameterizes the generation process of handwritten characters using delta-lognormal theory~\cite{motion-model2}. Although this approach can be used to express the generation process of handwriting in an interpretable way, it requires strong assumptions such as a simplification of the generation process. The stochastic model-based approach expresses the process of handwriting generation using a stochastic model, which allows for flexible modeling. In~\cite{bayesian-generation}, the strokes of the pen were generated stochastically using a Bayesian network. Graves used the MDN to stochastically generate and predict handwritten characters~\cite{1308.0850}. An attempt was also made to generate handwritten characters using a spiking neural network in~\cite{spiking-generation}.


\subsection{Prediction model considering bifurcation}
Several methods of bifurcation-aware predictions have also been proposed.
In~\cite{MDP_work1, MDP_work2}, the Markov decision process is used to predict human behavior considering the bifurcation. There are several attempts to predict time-series by considering class information and bifurcation simultaneously. Pool et al. proposed a prediction model for autonomous driving that can represent bifurcation by combining linear dynamical systems for each type of behavior of autonomous cars~\cite{mixture-LDS}. Deo and Trivedi proposed a model that classifies the trajectory patterns of automobiles into six classes and predicts trajectories according to the estimated class probabilities~\cite{uncertainty_work2, multi-modal-prediction2}. Tang and Salakhutdinov proposed a model for multimodal prediction without the explicitly labeling of tracking patterns~\cite{multiple-futures-prediction}.
Makansi et al. proposed a two-stage model based on the MDN, with hypothetical sampling by the Winner-Takes-All loss and distribution fitting, to avoid mode collapse~\cite{Makansi_2019_CVPR}. \par
Our trial can be differentiated from them at the following points. First, in our model, the prediction task is factorized into a class-wise coordinate prediction module and a single class probability module and trains both modules in an end-to-end framework (this implies that our model is solving a multi-task training task).
Second, compared to the tasks of the past attempts 
(e.g., car trajectory prediction), handwriting trajectories have very different from many aspects. 
Especially, we need to deal with nonstationarity because handwriting trajectories show different characteristics at each time point.
Third, our main focus is not only to get more accurate prediction but also to analyze the inter-class bifurcation structures underlying the specific target, i.e., handwritings. We therefore show many analysis results to understand the structures.


\section{Class-guided prediction}
\begin{figure}[t]
	\centering
    \includegraphics[width=\hsize]{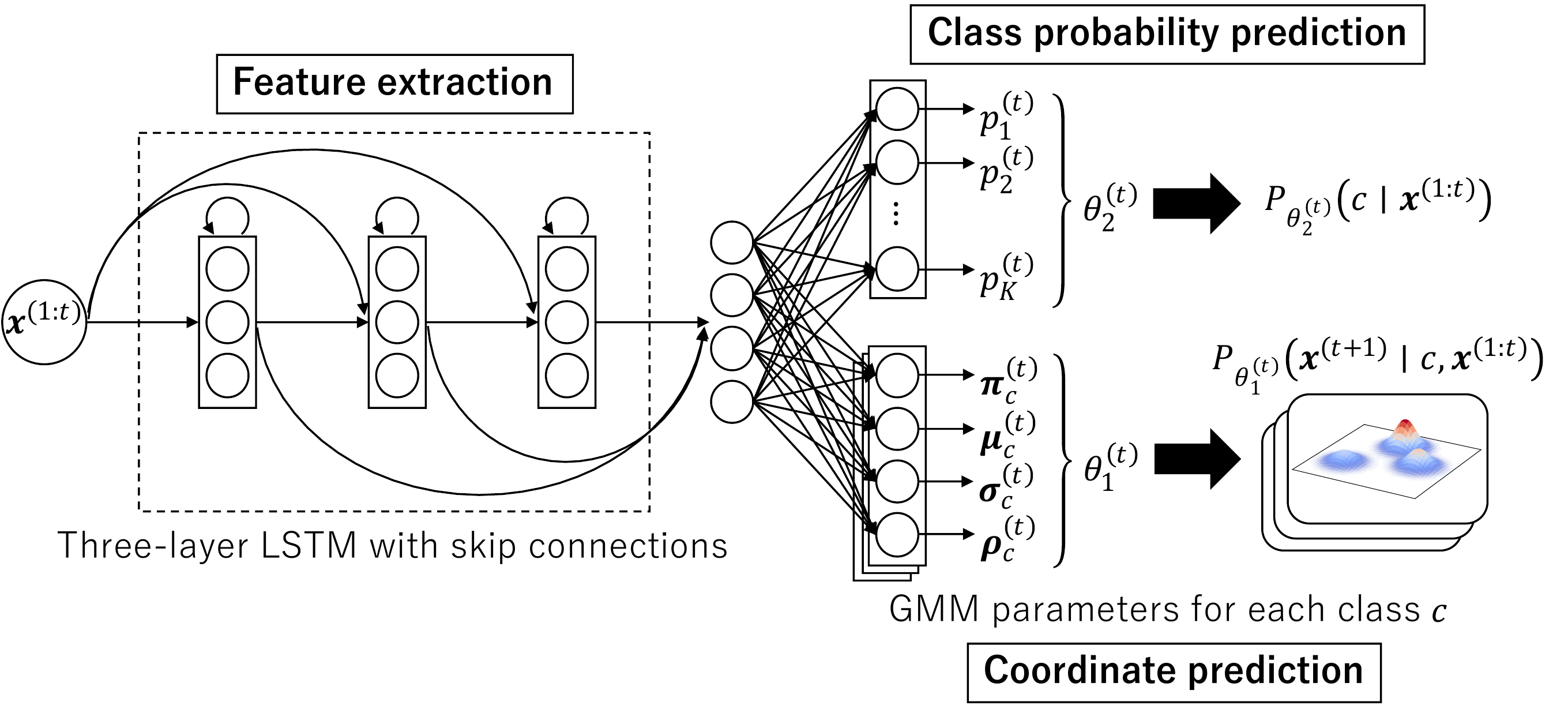}\\[-2mm]
    \caption{Structure of the proposed CGP model.} 
	\label{fig:CGP_structure}\vspace{-3mm}
\end{figure}
\subsection{Formulation of class-guided prediction}
The proposed CGP model provides the distribution of the next pen-tip coordinate $\bm{x}^{(t+1)}\in \mathbb{R}^2$ as its prediction result. Specifically, 
given a sequence $\bm{x}^{(1:t)}$, the CGP model provides the posterior distribution $P(\bm{x}^{(t+1)} \mid \bm{x}^{(1:t)})$. Instead of predicting a single pen-tip coordinate, the distribution estimation is suitable for dealing with bifurcations by providing multiple possible trajectories as random samples from the distribution. In addition, it is also useful to understand how the prediction uncertainty increases for a longer-term prediction.\par
The key idea of the proposed model is to factorize the distribution $P(\bm{x}^{(t+1)} \mid \bm{x}^{(1:t)})$ as the class-weighted sum of the class-conditional distribution of coordinates:
\begin{eqnarray}
\lefteqn{P(\bm{x}^{(t+1)} \!\mid\! \bm{x}^{(1:t)})}\nonumber\\ 
    &&=\! \sum_{c=1}^K P_{\theta_1^{(t)}}(\bm{x}^{(t+1)} \!\mid\! c, \bm{x}^{(1:t)})P_{\theta_2^{(t)}}(c \!\mid\! \bm{x}^{(1:t)}),
   \label{eq:factorize}
\end{eqnarray}
where $K$ is the number of classes and $\theta_1^{(t)}$ and $\theta_2^{(t)}$ are distribution parameters at $t$. Those parameters are time-variant because of the nonstationarity of handwriting trajectories.
\par
With this class-wise factorization, we can expect that the model acquires the inter-class bifurcation structure (like `0' and `6') more explicitly. If we try to model 
$P(\bm{x}^{(t+1)} \!\mid\! \bm{x}^{(1:t)})$ without class-wise factorization by, say, a single Gaussian mixture model such as MDN, there is a risk that the inter-class bifurcation parts are mixed-up into a single Gaussian component with a large covariance. For example, one of the Gaussian components for the ending parts of `0' and `6' might have a large covariance to cover both ending directions. In contrast, by the above factorization, each distribution $P_{\theta_1^{(t)}}(\bm{x}^{(t+1)} \!\mid\! c, \bm{x}^{(1:t)})$ is trained within each class and not disturbed by other classes. Therefore, we have $K$ sharper distributions with less overlaps---namely, we can catch the inter-class bifurcation structure more explicitly. 
%
\subsection{Prediction model by neural networks}
Fig.~\ref{fig:CGP_structure} shows the structure of the proposed CGP model. The CGP model consists of three network modules. The first module is prepared for 
feature extraction from $\bm{x}^{(1:t)}$ and composed of three layers of LSTM with skip connections between them. The skip connections mitigate the vanishing gradient problem and facilitate the learning of deep neural networks~\cite{1308.0850}. \par
The second and third modules are prepared for the factorization of (\ref{eq:factorize}). The second module is the coordinate prediction module for estimating $\theta_1^{(t)}$ and composed of a single fully-connected (FC) layer. The third module is the class probability prediction module for $\theta_2^{(t)}$
and also composed of a single FC layer. The 
approach that a neural network outputs the distribution parameters (of, especially, a mixture distribution) is inspired by the MDN~\cite{MDN}.\par
In this paper, a Gaussian mixture distribution is used to represent $P_{\theta_1^{(t)}}(\bm{x}^{(t+1)} \mid c, \bm{x}^{(1:t)})$. By preparing a mixture distribution for each class, its parameter set becomes
$$
\theta_1^{(t)}\!=\!\left\{\left(\pi^{(t)}_{c,m}, \bm{\mu}^{(t)}_{c,m}, \bm{\sigma}^{(t)}_{c,m}, \rho^{(t)}_{c,m}\right)
\!\mid\! m\!=\!1,\!\ldots,\!M, c\!=\!1,\!\ldots,\!K\right\},
$$
where $M$ is the number of Gaussian components per class, $\pi^{(t)}_{c,m}$ the mixture coefficient,  $\bm{\mu}^{(t)}_{c,m}$ is the mean,  $\bm{\sigma}^{(t)}_{c,m}$ is the standard deviation,
and $\rho^{(t)}_{c,m}$ is the correlation 
coefficient\footnote{The parameters $\bm{\sigma}^{(t)}_{c,m}=(\sigma_1, \sigma_2)$ and  $\rho^{(t)}_{c,m}$ specify the covariance matrix as
$\left(\begin{array}{cc}\sigma_1^2 &\rho^{(t)}_{c,m}\sigma_1\sigma_2 \\
\rho^{(t)}_{c,m}\sigma_1\sigma_2 & \sigma_2^2 \end{array}\right)$.}.
These parameters are estimated as the output of the coordinate prediction module\footnote{Several post-operations are applied to the network outputs to make them valid as the parameters of a probabilistic distribution. Specifically, the normalization to make $\sum_m\pi^{(t)}_{c,m}=1$, the range extension $\exp{\bm{\sigma}}^{(t)}_{c,m}\to \bm{\sigma}^{(t)}_{c,m}$,  and the range limitation 
$ \tanh{\rho^{(t)}_{c,m}}\to \rho^{(t)}_{c,m}$.
}  to the input $\bm{x}^{(1:t)}$.
\par
For the class probability $P_{\theta_2^{(t)}}(c \mid x^{(1:t)})$, a categorical distribution is used, whose parameter set is 
$$
 \theta_2^{(t)} = \{p^{(t)}_c \mid c=1,\ldots,K\}.
$$
Like $\theta_1^{(t)}$, these parameters are estimated as the outputs of the class probability prediction 
module\footnote{The outputs are also normalized so that $\sum_c p^{(t)}_c=1$.} to the input $\bm{x}^{(1:t)}$.\par

\subsection{Training the model}
The trainable weights of the CGP model, namely, the weights of LSTM layers and FC layers, are trained with a given training dataset. Given a set of $N$ training sequences and corresponding class 
labels $\{\bm{x}_n^{(1:T)}, \bm{y}_n\}_{n=1}^N$, where $\bm{y}_n = (y_{n,c})_{c=1,\ldots,K}$ is the class label encoded as a one-hot vector, the CGP model is trained by minimizing the following loss function $L$: 
$$
    L = \sum_{n=1}^N \sum_{t=1}^T  L_{\rm coord}^{(n,t)} + L_{\rm class}^{(n,t)},
$$
where $L_{\rm coord}^{(n,t)}$ is the negative log-likelihood loss for the coordinate prediction module and evaluates the likelihood of $\bm{x}^{(t+1)}$ by the current mixture distribution:
$$
    L_{\rm coord}^{(n,t)} \!=\! -\log \! \sum_{c=1}^K \! y_{n,c} \! \sum_{j=1}^M \!  \pi^{(t)}_{c,j}\mathcal{N}\left(\bm{x}^{(t+1)}|\bm{\mu}^{(t)}_{c,j}, \bm{\sigma}^{(t)}_{c,j}, \rho^{(t)}_{c,j}\right), 
$$
and $L_{\rm class}^{(n,t)}$ is the the cross entropy loss for the class probability prediction module and evaluates how the module outputs correct class probabilities: 
$$
    L_{\rm class}^{(n,t)} = -\sum_{c=1}^K y_{n,c} \log p^{(t)}_c.
$$
Since the entire model is formulated using only differentiable operations, its weights can be updated through backpropagation in an end-to-end manner. \par
%
\subsection{Stochastic prediction by the model\label{sec:predict}}
A two-step sampling procedure is employed to obtain the coordinate at $t+1$. In the first step, a class $\tilde{c}$ at $t$ is sampled according to the class probability distribution $P_{\theta_2^{(t)}}(c \!\mid\! \bm{x}^{(1:t)})$. The predicted coordinate $\bm{x}^{(t+1)}$ is sampled from $P_{\theta_1^{(t)}}(\bm{x}^{(t+1)} \!\mid\! \tilde{c}, \bm{x}^{(1:t)})$.
For predicting the coordinate at $t+\Delta t$, this two-step sampling procedure is repeated $\Delta t - 1$ times by concatenating the predicted results and $\bm{x}^{(1:t)}$.

\section{Experiment}
\subsection{Experimental setups}
To evaluate the validity of the CGP, we conducted an experiment to predict handwriting. We used the UNIPEN database (Train-R01/V07, 1a)~\cite{unipen}, which contains online handwritten digits as sequences of two-dimensional coordinates. We normalized these sequences in the range [0, 127] for each dimension and the time length of 50. Instead of directly inputting the coordinate sequences into the model, we converted each coordinate into a relative coordinate that is the difference from the coordinate one time prior to the current time point. Since sequences of relative coordinates include the relationships of two temporally adjacent coordinates to each other, the model is expected to catch movements of the handwriting more easily than when absolute coordinate sequences are used. This dataset is slightly class-imbalanced and contains approximately 1,300 sequences for each class. We randomly divided the dataset into 70\% for the training set, 10\% for the validation set, and 20\% for the test set.　

For comparison, we used deterministic prediction using LSTMs (D-LSTM), a model combining LSTMs and an MDN (hereafter referred to simply as the MDN), and the 1-nearest neighbor (1-NN). In the CGP model and the MDN, which output mixture distributions, the total number of components was unified to 40, i.e., four components were assigned to each digit class in the CGP model. The numbers of training epochs for the D-LSTM, MDN, and CGP were determined based on the criterion that the validation loss did not decrease for 10 epochs. We used the mean squared error for the D-LSTM as the loss function.

In the MDN, the predicted trajectory at $t+1$ was obtained by sampling $x^{(t+1)}$ from $P(x^{(t+1)} \mid x^{(1:t)})$. As in the CGP model, the trajectory at $t+\Delta t$ ($\Delta t \geq 2 $) was predicted by the process of Section~\ref{sec:predict}.

\subsection{Qualitative evaluation}
We investigated the characteristics of the CGP model by qualitatively evaluating its representative results. Fig.~\ref{fig:prediction_example} shows an example of the predictions made by the CGP model. 
\begin{figure*}[t]
	\centering
        \includegraphics[width=0.087\hsize]{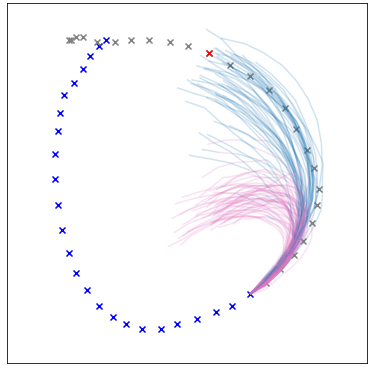}
        \includegraphics[width=0.087\hsize]{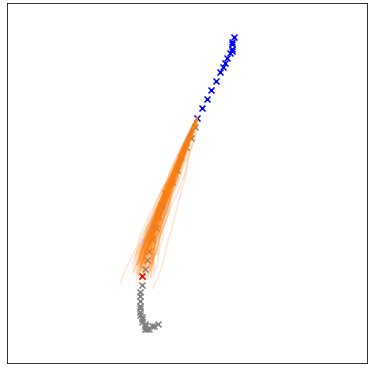}
        \includegraphics[width=0.087\hsize]{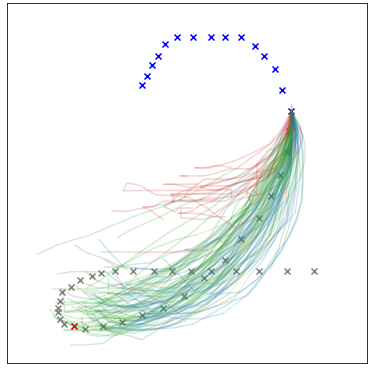}
        \includegraphics[width=0.087\hsize]{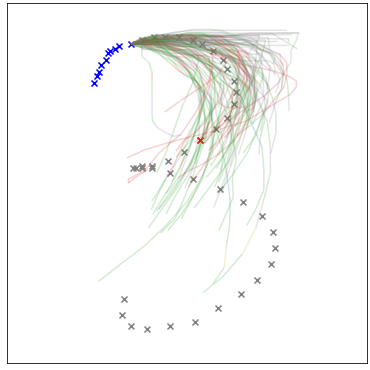}
        \includegraphics[width=0.087\hsize]{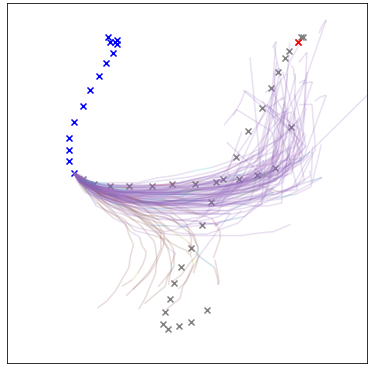}
        \includegraphics[width=0.087\hsize]{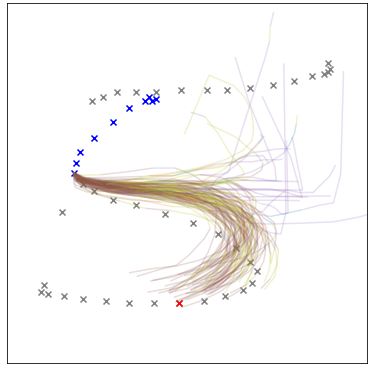}
        \includegraphics[width=0.087\hsize]{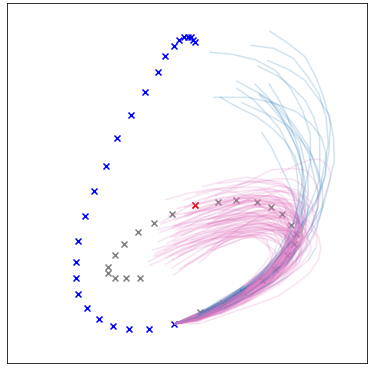}
        \includegraphics[width=0.087\hsize]{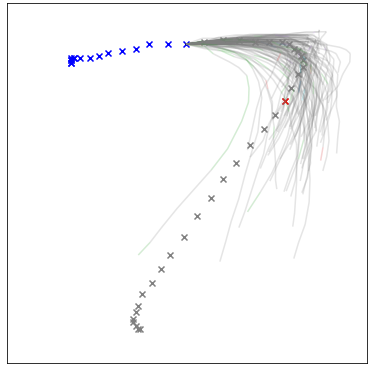}
        \includegraphics[width=0.087\hsize]{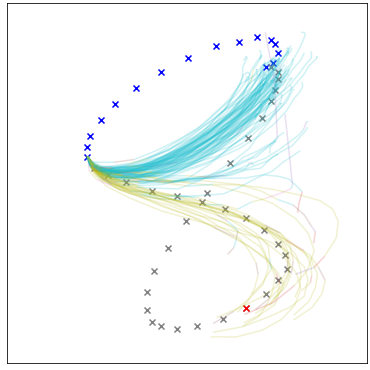}
        \includegraphics[width=0.087\hsize]{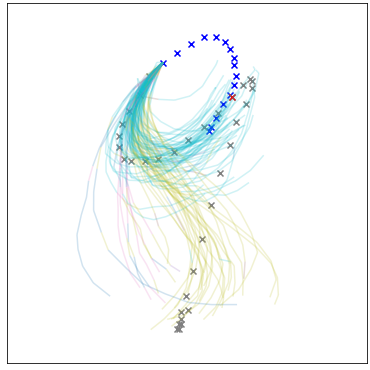}
        \includegraphics[height=0.1\hsize]{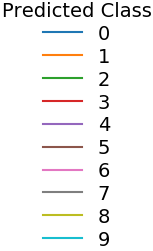}
	\caption{Example of the predictions made by the CGP model. The blue dots are the inputted coordinate sequences, and the gray dots are subsequent true coordinate sequences. The red dots represent the final time point of the coordinate sequences. The colored lines are the class-conditional trajectories predicted by the CGP model.}
	\label{fig:prediction_example}
\end{figure*}
Note that this figure was drawn by overlying results of 100 samplings, and the subsequent figures are drawn in the same way. These examples confirmed that the CGP model can predict the trajectories of multiple classes. For example, in the panel at the top-left corner, most of the predicted trajectories branch to classes `0' and `6' because there is still the possibility of being in either class at this time point.

Figs.~\ref{fig:bifurcation_t_class2and3}(a) and (b) show changes in the predictions made by the CGP model when the length of the input was increased. 
\begin{figure}[t]
    \centering
	\begin{minipage}{\hsize}
	    \centering
        \includegraphics[width=0.24\hsize]{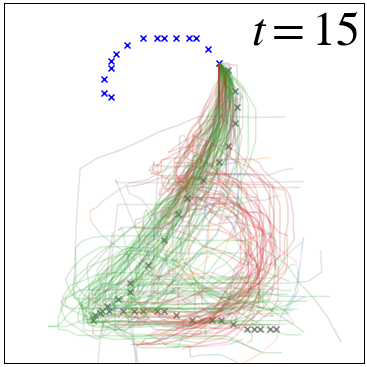}
        \includegraphics[width=0.24\hsize]{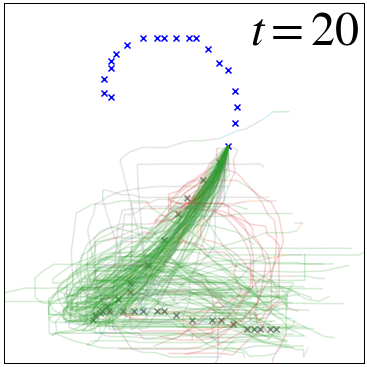}
        \includegraphics[width=0.24\hsize]{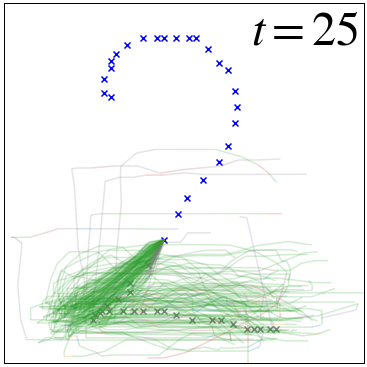}
        \includegraphics[width=0.24\hsize]{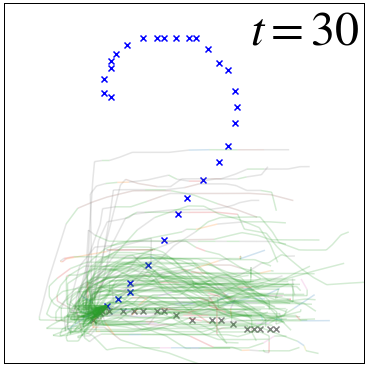}\\[-2mm]
	    \subcaption{Correct class: `2'}
	    \label{fig:bifurcation_t_class2}
	\end{minipage} \\
	\vspace{4mm}
	\begin{minipage}{\hsize}
	    \centering
        \includegraphics[width=0.24\hsize]{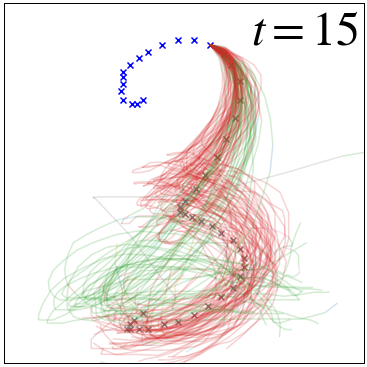}
        \includegraphics[width=0.24\hsize]{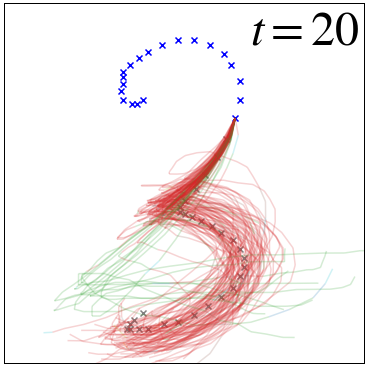}
        \includegraphics[width=0.24\hsize]{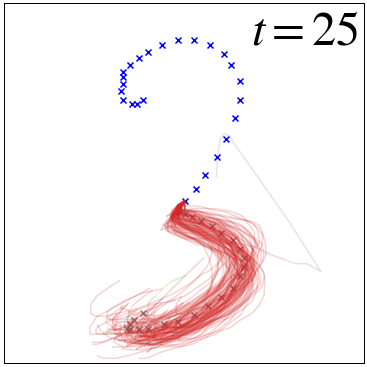}
        \includegraphics[width=0.24\hsize]{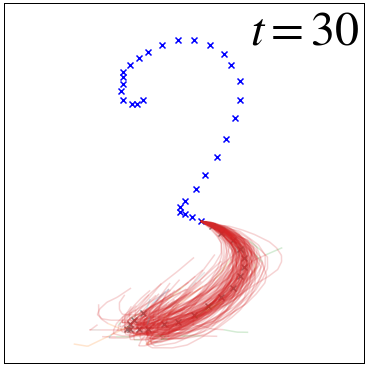}\\[-2mm]
	    \subcaption{Correct class: `3'}
	    \label{fig:bifurcation_t_class3}
	\end{minipage} \\
	\caption{Predictions for `2' and `3' while increasing input length $t$. The colors represent the same items as in Fig.~\ref{fig:prediction_example}.}
	\label{fig:bifurcation_t_class2and3}
\end{figure}
In both cases, the predictions contained the possibility of the bifurcation of either `2' or `3' in the early stages. As the length of the input increases, the number of predictions for the correct class increases in both cases. Therefore, the CGP model made predictions based on an ambiguous class prediction in the early stages, and the certainty of its predictions increased with the input length. This result is reasonable because the longer the input length was, the more confident the prediction was.

Fig.~\ref{fig:bifurcation_dt_class5} shows an example of a prediction by the CGP model with inputs up to $t = 20$ while varying $\Delta t$, where the pen-up for `5' was performed. 
\begin{figure}[t]
	\centering
	\includegraphics[width=0.24\hsize]{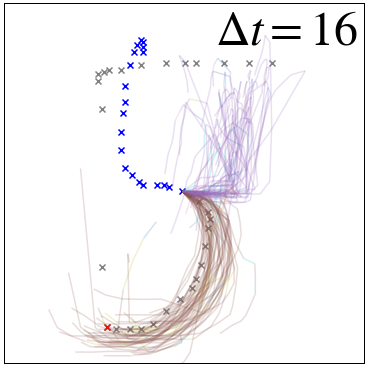}
	\includegraphics[width=0.24\hsize]{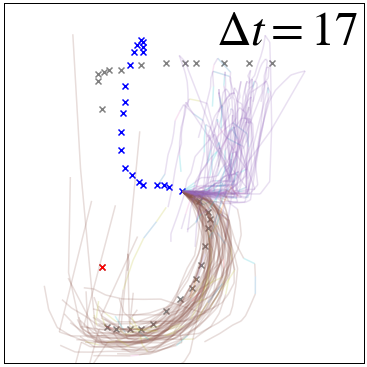}
	\includegraphics[width=0.24\hsize]{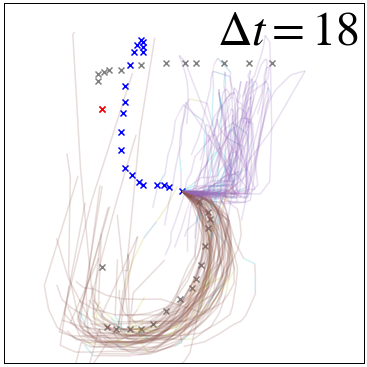}
	\includegraphics[width=0.24\hsize]{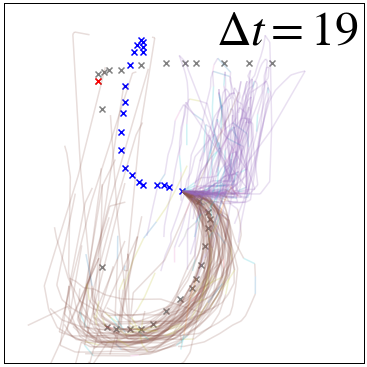}
	\caption{Trajectories predicted by the CGP model for `5' while varying $\Delta t$ with fixed inputs up to $t = 20$. The colored lines have the same meanings as in Fig.~\ref{fig:prediction_example}.}
	\label{fig:bifurcation_dt_class5}
\end{figure}
Although accurate prediction is difficult because the stroke moves drastically when pen-up occurs, it was confirmed that the pen-up was predicted at an approximately accurate timing. Furthermore, the destination after the pen-up was a natural starting position for the subsequent stroke. Incidentally, this example contains strokes for `4' in the early stages of the prediction, and even in this case, the pen-up for `4' was correctly predicted.

Fig.~\ref{fig:comparison_of_long_term_prediction} shows a comparison of the results predicted by the CGP model, MDN, D-LSTM, and 1-NN. 
\begin{figure}[!t]
    \centering
	\begin{minipage}{0.24\hsize}
	    \centering
	    \includegraphics[width=1\hsize]{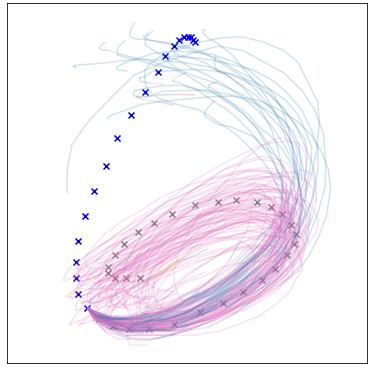}
	\end{minipage}
	\begin{minipage}{0.24\hsize}
	    \centering
	    \includegraphics[width=1\hsize]{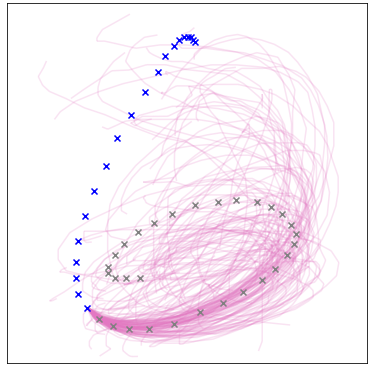}
	\end{minipage}
	\begin{minipage}{0.24\hsize}
	    \centering
	    \includegraphics[width=1\hsize]{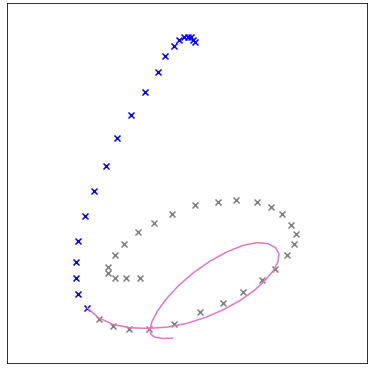}
	\end{minipage}
	\begin{minipage}{0.24\hsize}
	    \centering
	    \includegraphics[width=1\hsize]{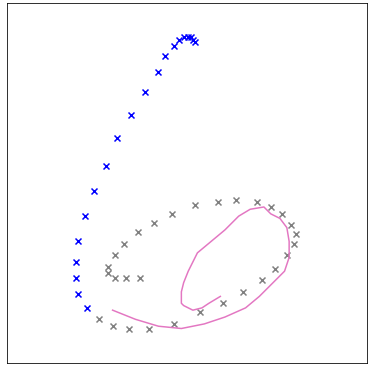}
	\end{minipage} \\
	\begin{minipage}{0.24\hsize}
	    \centering
	    \includegraphics[width=1\hsize]{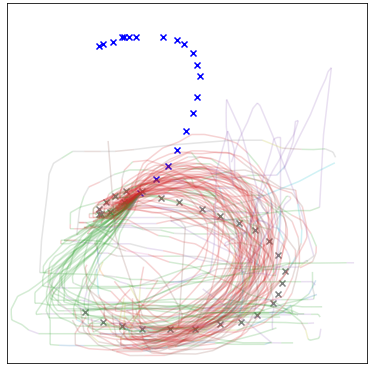}\\[-2mm]
	    \subcaption{CGP}
	    \label{fig:chapter3:long:proposed}
	\end{minipage}
	\begin{minipage}{0.24\hsize}
	    \centering
	    \includegraphics[width=1\hsize]{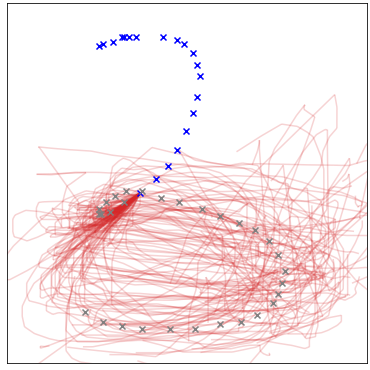}\\[-2mm]
	    \subcaption{MDN}
	    \label{fig:chapter3:long:MDN}
	\end{minipage}
	\begin{minipage}{0.24\hsize}
	    \centering
	    \includegraphics[width=1\hsize]{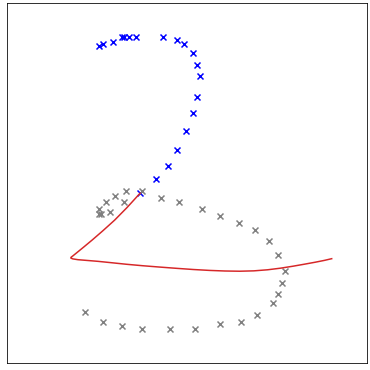}\\[-2mm]
	    \subcaption{D-LSTM}
	    \label{fig:chapter3:long:D-LSTM}
	\end{minipage}
	\begin{minipage}{0.24\hsize}
	    \centering
	    \includegraphics[width=1\hsize]{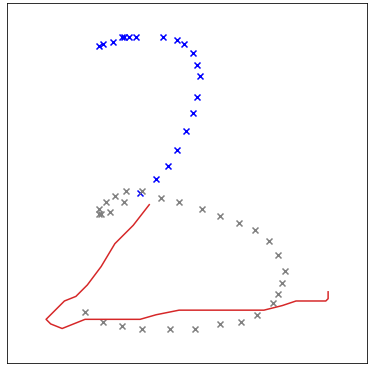}\\[-2mm]
	    \subcaption{1-NN}
	    \label{fig:chapter3:long:k-NN}
	\end{minipage} \\
	\caption{Comparison of the results of prediction by the CGP model, MDN, D-LSTM, and 1-NN. The figures in the top row show predictions for `6' and those in the bottom row show those for `3.'}
	\label{fig:comparison_of_long_term_prediction}
\end{figure}
Since the MDN is a probabilistic model as well as the CGP, the results of 100 samplings are drawn. The spread of the predicted lines is proportional to the variance of the predicted distribution. In the prediction for `6' by the CGP model (top of Fig. \ref{fig:chapter3:long:proposed}), a branch to `0' and `6' is apparent. The lines predicted by the MDN for `0' and `6' were not clearly separated, and it appeared to have been a unimodal prediction, rather than a bimodal prediction, with a large variance. This is because the role of each component was not clearly specified in the MDN. Therefore, the distribution predicted the MDN was ambiguous, having a large variance with the mean between `0' and `6.' In Figs.~\ref{fig:chapter3:long:D-LSTM} and \ref{fig:chapter3:long:k-NN}, the predictions of the D-LSTM and 1-NN deviate from the true values. In particular, for the prediction of `3' (bottom of Figs.~\ref{fig:chapter3:long:D-LSTM} and \ref{fig:chapter3:long:k-NN}), it seems that the D-LSTM and 1-NN recognized the class as `2' and subsequently predicted the corresponding trajectory. Since the D-LSTM and 1-NN can only make deterministic predictions, their performance worsened if a class was incorrectly predicted.

\subsection{Quantitative evaluation}
We quantitatively evaluated the performance of the CGP model. We used two criteria for evaluation: the root mean squared error (RMSE) and negative log-likelihood (NLL). 
We sampled 20 times from the predictive distribution for each test data and calculated both criteria.

In the evaluation based on the RMSE, we used two types of class-conditional RMSEs (RMSE 2 and RMSE 3), in addition to the ordinary RMSE (RMSE 1). This is because it is inadvisable to calculate errors for all predictions at a given time point in case there is a possibility of branching. For example, as shown in Fig.~\ref{fig:comparison_of_long_term_prediction}, when there is a possibility of predicting both the correct class `6' and the incorrect class `0', it is preferable to calculate the error only for the correct class.
We defined RMSE 2 as the RMSE for the sampling results from the majority class distribution, whereas we defined RMSE 3 as the RMSE from the sampling results of the correct class distribution.
In order to compute RMSE 2 and RMSE 3, we sampled coordinates from the majority class distribution for RMSE 2 and from the correct class distribution for RMSE~3.

\begin{figure}[t]
	\centering
	\begin{minipage}{0.49\hsize}
	    \centering
    	\includegraphics[width=1\hsize]{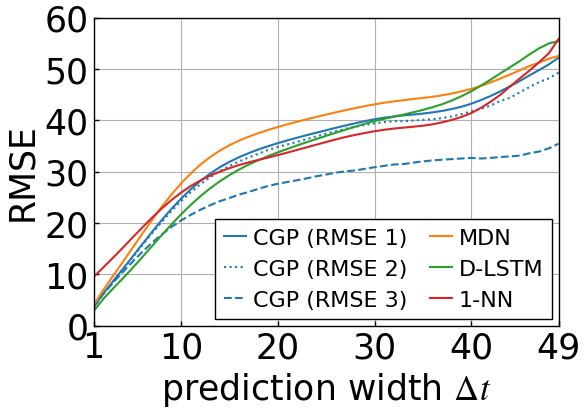}
    	\subcaption{RMSE}
    	\label{fig:rmse_all}
	\end{minipage}
	\begin{minipage}{0.49\hsize}
    	\centering
    	\includegraphics[width=1\hsize]{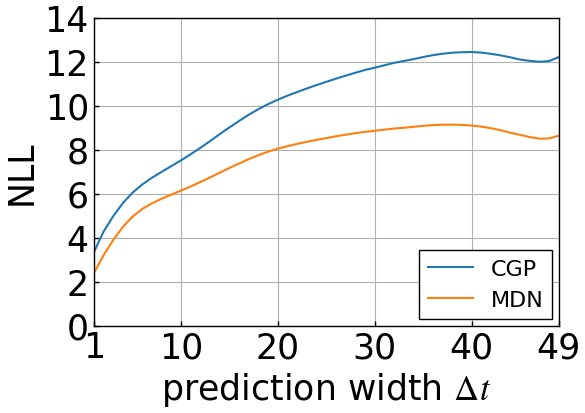}
    	\subcaption{Negative log-likelihood}
    	\label{fig:nll}
    \end{minipage}
    \caption{Performance of each model.}
    \vspace{-3mm}
\end{figure}
Fig.~\ref{fig:rmse_all} shows the RMSE values for each model. There is no significant difference between the errors of the CGP model and the other models in terms of RMSE 1. When compared in terms of RMSE 2, the error of the CGP model is slightly lower than those of the others for large values of $\Delta t$. When compared in terms of RMSE 3, the error of the CGP following $\Delta t = 9$ was remarkably lower than those of the other models. These results suggest that the CGP model can make more accurate predictions when it can correctly predict the class of the input time series.

\begin{figure*}[t]
    \centering
    \begin{minipage}{0.912\hsize}
        \includegraphics[width=0.195\hsize]{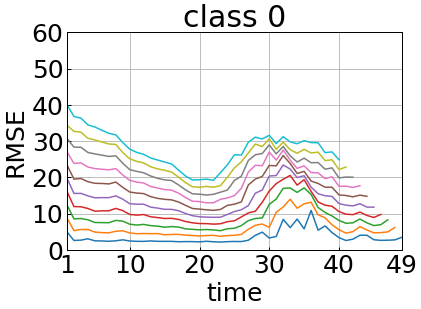}
        \includegraphics[width=0.195\hsize]{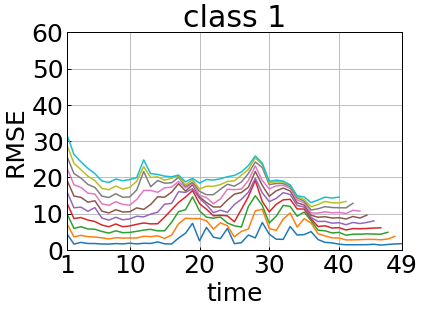}
        \includegraphics[width=0.195\hsize]{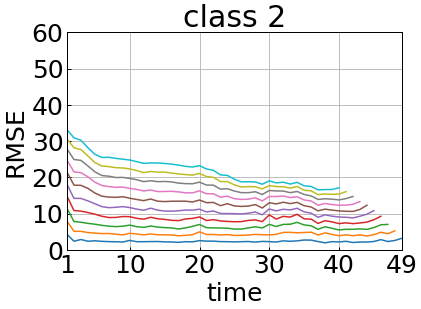}
        \includegraphics[width=0.195\hsize]{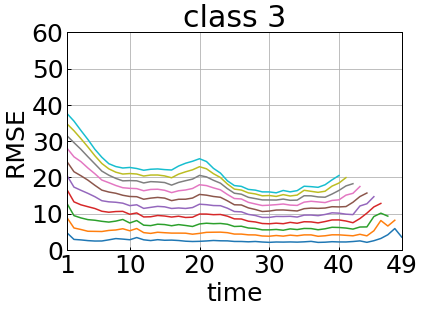}
        \includegraphics[width=0.195\hsize]{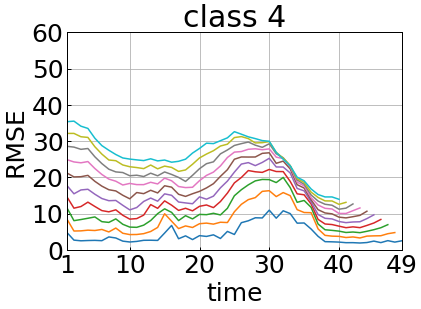}
        \\
        \includegraphics[width=0.195\hsize]{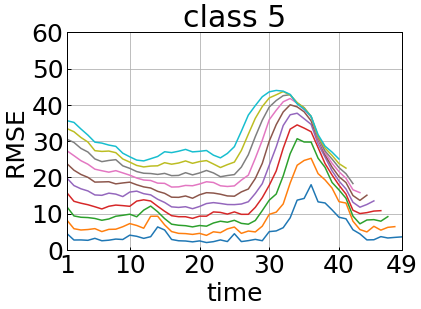}
        \includegraphics[width=0.195\hsize]{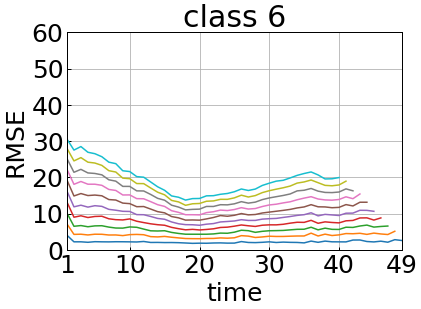}
        \includegraphics[width=0.195\hsize]{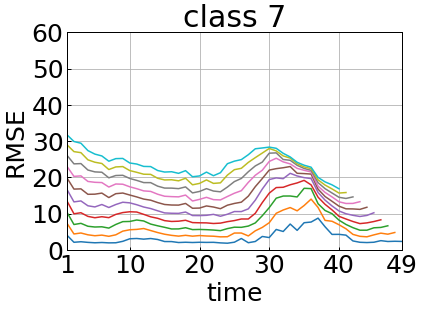}
        \includegraphics[width=0.195\hsize]{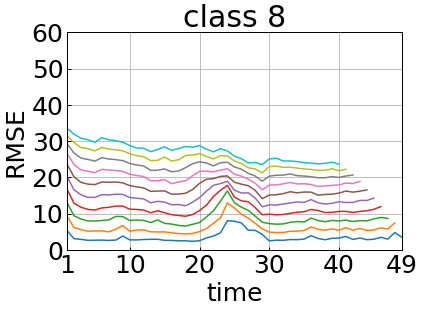}
        \includegraphics[width=0.195\hsize]{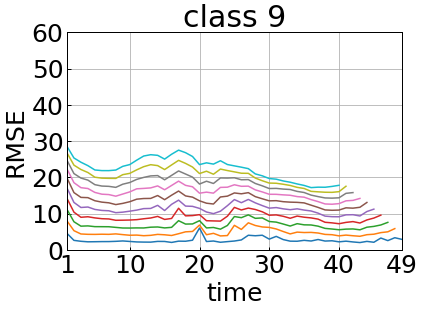}
    \end{minipage}
    \begin{minipage}{0.08\hsize}
        \includegraphics[width=1\hsize]{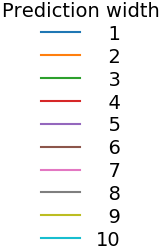}
    \end{minipage}\\[-2mm]
    \caption{Changes in RMSE for each class.}
    \label{fig:class-rmse}\vspace{-2mm}
\end{figure*}

\begin{figure*}[t]
    \centering
    \begin{minipage}{0.912\hsize}
        \includegraphics[width=0.195\hsize]{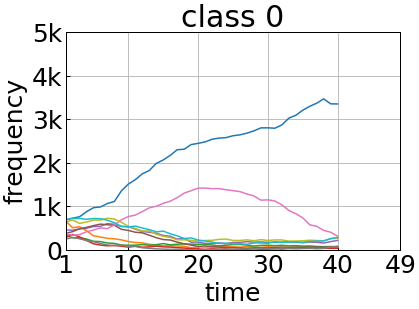}
        \includegraphics[width=0.195\hsize]{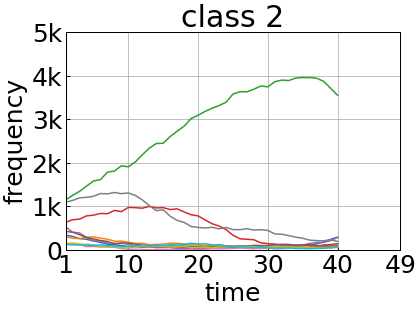}
        \includegraphics[width=0.195\hsize]{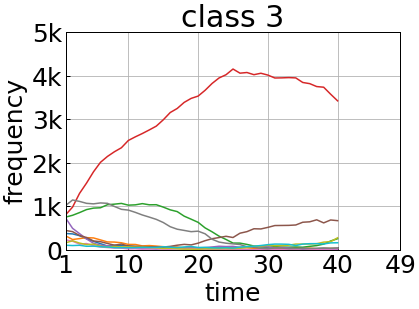}
        \includegraphics[width=0.195\hsize]{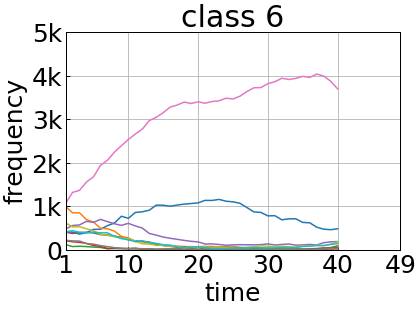}
        \includegraphics[width=0.195\hsize]{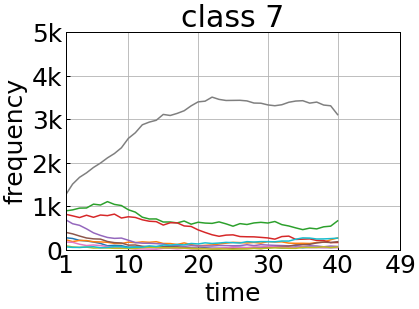}
    \end{minipage}\\[-2mm]
    \caption{Frequency of the selected classes in sampling ($\Delta t = 10$). The colored lines have the same meanings as in Fig.~\ref{fig:prediction_example}.}
    \label{fig:class-frequency}
    \vspace{-3mm}
\end{figure*}

Fig.~\ref{fig:nll} shows the results of evaluation using the NLL. The log-likelihood calculates how well the predicted distribution fits the real values, and its negative value is often used to assess density estimation: The smaller the NLL is, the better the distribution is. 

In Fig.~\ref{fig:nll}, the MDN shows a smaller NLL than the CGP model. As $\Delta t$ increased, the gap became more prominent. These results confirmed that the MDN generated the predicted distribution that fits the real values, whereas the CGP model also demonstrated competitive results. A possible explanation of this is the difference in the variance of distributions for long-term predictions. As shown in Fig.~\ref{fig:comparison_of_long_term_prediction}, the MDN tended to make a prediction with a larger variance than the CGP model. If the variance of each component is large, the likelihood tends to become large even if the mean of each component apart from the true value. For these reasons, we considered that the MDN ostensibly showed better NLL values despite ambiguous predictions.

\subsection{Bifurcation structure of characters}
Fig.~\ref{fig:class-rmse} shows changes in RMSE 1 according to the input time $t$ for each class and each prediction width $\Delta t$. Roughly speaking, the RMSE tended to decrease over time for all classes. However, for some classes, the time variation of the RMSE had a specific pattern rather than a monotonic decrease. For example, in class `5,' the RMSE increased rapidly around $t = 25$ and then decreased around $t = 35$. Moreover, this class-specific pattern was the same regardless of the predicted width.

This pattern of time variation in the RMSE for each class is related to the bifurcation in the handwriting prediction. For example, `0' and `6' had a similar trajectory until the middle stage of the writing process, but bifurcate after that. This resulted in a decrease in the RMSE up to the bifurcation point of the character (around $t=18$) because a common prediction could have been made. However, the RMSE increased because it was not possible to determine whether the class was `0' or `6' when predicting after the bifurcation point. As the input series became longer and exceeded $t=35$, the prediction of the character class became clearer, resulting in a decrease in the RMSE. That is, there were similarities in the time variation of the RMSE among classes that have common parts in the handwriting trajectory, which was influenced by the bifurcation structure of the characters.

The frequency of sampled class labels according to the input time $t$ is shown in Fig.~\ref{fig:class-frequency}. Note that the figures are drawn for each correct class with the prediction width fixed to $\Delta t = 10$, and that a particularly confusing pair (`0' and `6') and triplet (`2,' `3,' and `7') were selected. For all cases, the frequency of multiple classes was high in the early stage of the prediction, which means that the class predictions were ambiguous. Until the middle stage of prediction, the frequency of classes with a common part in the trajectory, e.g., `0' and `6,' increased, while the frequencies of the other classes decreased. This indicates that the number of possible character classes of the input series was limited as the time length of the input increased. The state where the frequency of these few classes was high persisted because it was not possible to determine the class until the bifurcation point. Once the input time had exceeded $t=23$, the frequency of incorrect classes tended to decrease. The classes `0' and `6' in Fig.~\ref{fig:class-rmse} show that the RMSE decreased at around $t = 35$, which is consistent with the result of the selected class frequency. The graphs of classes `2,' `3,' and `7' in Fig.~\ref{fig:class-frequency} show that the prediction had three classes of possibilities, which means that `2,' `3,' and `7' each had a three-way bifurcated structures.

\section{Conclusion}
In this paper, we proposed a temporal prediction model that can handle the bifurcation structures related to class information. By combining class prediction and class-conditional coordinate prediction, the proposed class-guided prediction (CGP) model learns the bifurcation structure explicitly. In experiments using the UNIPEN online handwritten character dataset, we verified that the CGP model can represent and handle the bifurcation structure of handwritten characters, and can predict their trajectories with a smaller error.

In future work, we will utilize the learned bifurcation structure for several applications and scientific investigations. For example, we can realize early classification by knowing that no bifurcation after the current time point~\cite{early1,early2}. It is also possible to combine our framework with other temporal pattern generation models. In fact, the ability of predicting the future uncertainty will be useful for the reinforcement learning-based generative models, since those models rely on the future reward prediction. 

} 
\section*{Acknowledgment}
This work was partially supported by JSPS KAKENHI Grant Number JP17H06100 and JST ACT-I Grant Number JPMJPR18UO.

\bibliographystyle{IEEEtran}
\bibliography{refer}

\end{document}